\documentclass[10pt, runningheads]{llncs}
\usepackage{graphicx}
\usepackage{booktabs} 
\usepackage[numbers]{natbib}
\usepackage{amsmath}
\usepackage{amsfonts}
\usepackage{subcaption}
\usepackage{tabularx}
\usepackage{floatrow}
\usepackage{blindtext}
\usepackage[font=small,labelfont=bf]{caption}

\newfloatcommand{capbtabbox}{table}[][\FBwidth]
\usepackage{color}
\newcommand{\eat}[1]{}

\makeatletter
\renewcommand\section{\@startsection{section}{1}{\z@}%
	{-8\p@ \@plus -4\p@ \@minus -4\p@}%
	{6\p@ \@plus 4\p@ \@minus 4\p@}%
	{\normalfont\large\bfseries\boldmath
		\rightskip=\z@ \@plus 8em\pretolerance=10000 }}
\renewcommand\subsection{\@startsection{subsection}{2}{\z@}%
	{-8\p@ \@plus -4\p@ \@minus -4\p@}%
	{6\p@ \@plus 4\p@ \@minus 4\p@}%
	{\normalfont\normalsize\bfseries\boldmath
		\rightskip=\z@ \@plus 8em\pretolerance=10000 }}
\renewcommand\subsubsection{\@startsection{subsubsection}{3}{\z@}%
	{-4\p@ \@plus -4\p@ \@minus -4\p@}%
	{-1.5em \@plus -0.22em \@minus -0.1em}%
	{\normalfont\normalsize\bfseries\boldmath}}
\makeatother

\begin{document}
	\title{Enhanced Modality Transition\\ for Image Captioning}
	\author{Ziwei Wang\orcidID{0000-0002-0107-7347},
		Yadan Luo\orcidID{0000-0001-6272-2971},\\
		and  Zi Huang\orcidID{0000-0002-9738-4949}}
	\authorrunning{Z. Wang et al.}
	%
	\institute{School of Information Techonology and Electrical Engineering,\\
		The University of Queensland, Brisbane, Australia\\
		\email{ziwei.nj@gmail.com}, \email{lyadanluol@gmail.com}, \email{huang@itee.uq.edu.au}}
	
	\maketitle              
	\begin{abstract}
		Image captioning model is a cross-modality knowledge discovery task, which targets at automatically describing an image with an informative and coherent sentence. To generate the captions, the previous encoder-decoder frameworks directly forward the visual vectors to the recurrent language model, forcing the recurrent units to generate a sentence based on the visual features.
		Although these sentences are generally readable, they still suffer from the lack of details and highlights, due to the fact that the substantial gap between the image and text modalities is not sufficiently addressed. 
		In this work, we explicitly build a Modality Transition Module (MTM) to transfer visual features into semantic representations before forwarding them to the language model. During the training phase, the modality transition network is optimised by the proposed modality loss, which compares the generated preliminary textual encodings with the target sentence vectors from a pre-trained text auto-encoder.
		In this way, the visual vectors are transited into the textual subspace for more contextual and precise language generation.
		The novel MTM can be incorporated into most of the existing methods. Extensive experiments have been conducted on the MS-COCO dataset demonstrating the effectiveness of the proposed framework, improving the performance by 3.4\% comparing to the state-of-the-arts.
		
		\keywords{Image Captioning  \and Cross-modal \and Modality Transition}
	\end{abstract}

	\section{Introduction} \label{sec:introduction}
	Vision and language are important unstructured data sources in knowledge discovery research. In recent years, the intelligent multimodal content understanding models have been extensively proposed and studied to meet the increasing demand for multimedia knowledge management.
	With advancements of deep neural networks, a number of deep content understanding models are beginning to sprout up in various fields including image captioning, video captioning, and visual question answering (VQA). These well-trained models can be further embedded in a larger multimedia database for the down-streaming tasks of efficient retrieval, knowledge discovery, etc.
	\eat{The performance of these models, such as cross-modal information retrieval, multi-modal fusion and image-text matching, have been significantly improved with the advancements of deep neural networks.}
	\eat{Majority of the intelligent multimodal systems exploit deep convolutional neural networks (CNN) to extract visual features, and recurrent neural networks (RNN) to represent or generate word sequences.
	The deep representations can be utilised for natural language descriptions in image captioning, or for answer generating in VQA.}
	
\begin{figure}[!t]
		\centering
		\subfloat[Baseline. \label{fig:overview_baseline}]{%
			\includegraphics[trim=0cm 3.6cm 0cm 3.4cm, width=0.94\textwidth, height = 2.4cm, clip=false]{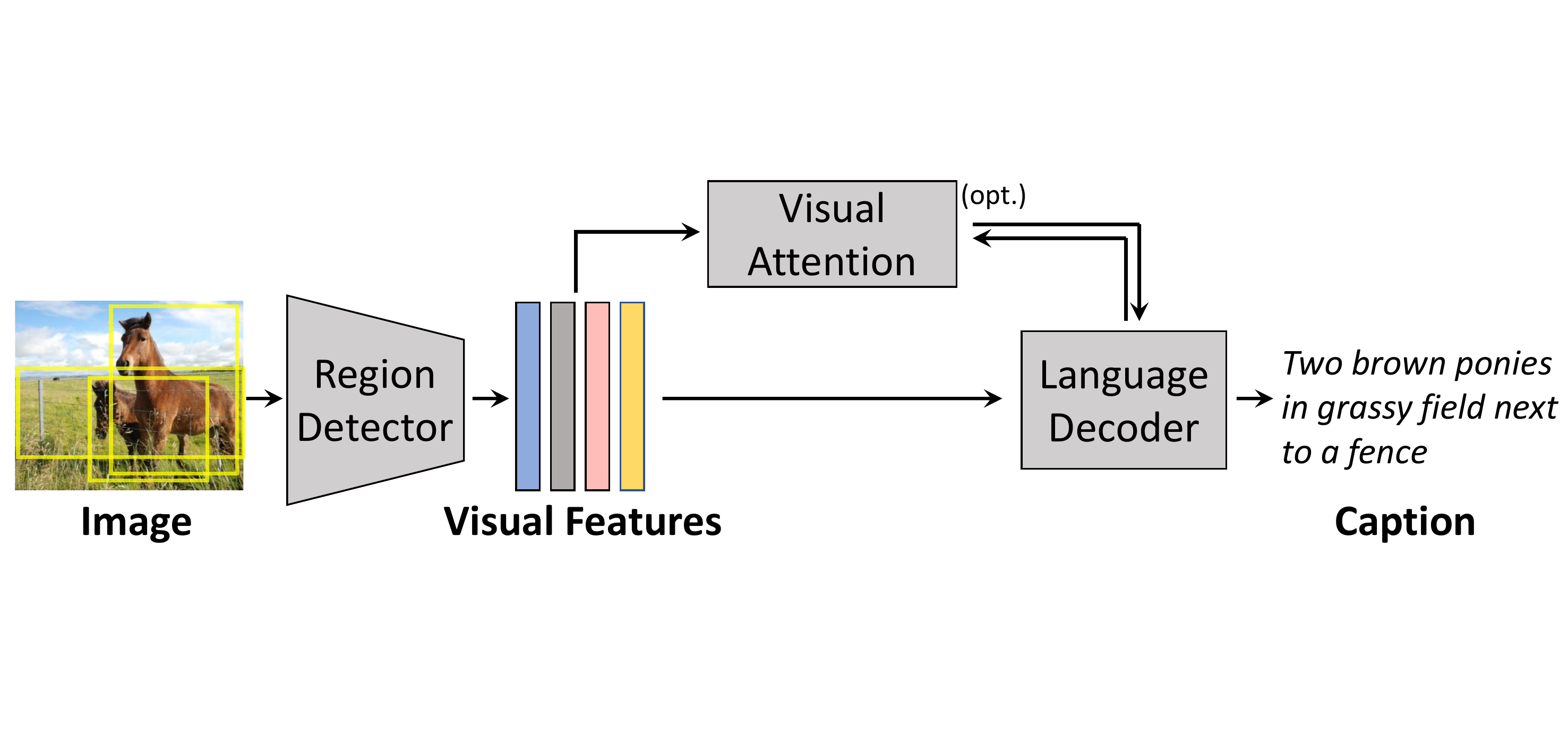}%
		}
		\\
		\vspace{0.4cm}
		\subfloat[Modality Transition.\label{fig:overview_ours}]{%
			\includegraphics[trim=0cm 0.8cm 0cm 0.8cm, width=0.94\textwidth,height = 4.2cm, clip=false]{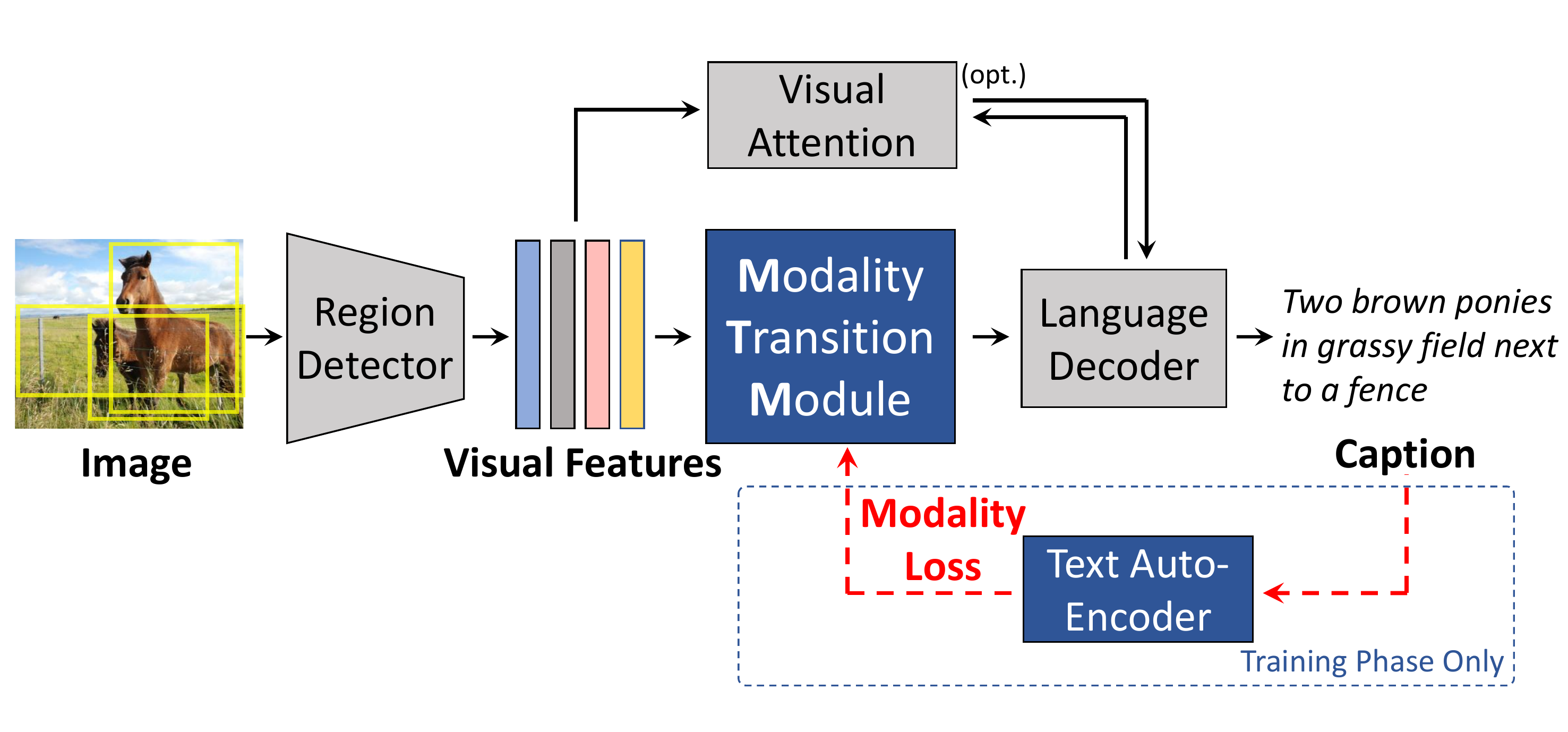}%
		}
		
		\caption{Framework overview. (a) The conventional image captioning baseline extracts region features and directly decodes the visual vectors into the caption via language decoder, optionally equipped with visual attention module. (b) The proposed modality transition framework explicitly transfers the vision features into global textual features with the proposed Modality Transition Module (MTM). The modality loss is calculated by aligning MTM outputs with the auto-encoded caption embeddings.}
		\label{fig:overview}
		
	\end{figure}

	In particular, this paper focuses on image captioning, which engages vision and language in a concise way.
	Given an image, the vision encoder extracts the visual features from the RGB picture, then a language model decodes the visual representations into a sentence-level description, as illustrated in Fig.\ref{fig:overview_baseline}.
	Typically, for the vision encoder, the CNN extracts visual feature maps to represent the visual content, followed by an optional region detector (e.g. Faster-RCNN) for recognising object-wise regions and outputting  the refined regional features.
	For the language decoder, the visual features initialise the language model as an input to generate the word one-by-one given the first ``visual'' word embeddings using Long Short-Term Memory (LSTM)~\cite{lstm}.
	Other techniques have been made to further improve the caption quality, such as incorporating visual attention, attributes, scene graphs, and reinforcement learning.
	
	\eat{Furthermore, a number of modules are proposed to explicitly extract salient objects or areas from the image to facilitate sentence generation such as visual attention mechanism. When the language model is decoding the words, the visual attention adaptively shifts the sentence focus on the most relevant image regions given the current generated word.}
	\eat{Another concurrent model extracts attributes from the visual images using multi-instances learning framework, so the language decoder can emphasise on the related attributes.}
	
	Most of the work is built on such an encoder-decoder framework, based on the assumption that visual features can be perfectly recognised by language decoder. The vision and language modalities are assumed to be interchangeable by inserting the global visual features as the first visual ``word'' for the language model.
	However, this assumption overlooks the underlying facts that the visual features and word embeddings are still trained in different ways despite that efforts have been made during the fine-tuning. The visual feature encoder is generally pre-trained to classify 1,000 categories (e.g. ImageNet) given the image, whilst the language model is trained to predict the next word from a vocabulary with approximately 10,000 words. Therefore, the intrinsic modality gap between vision and language still exists, and the previous methods made limited efforts to address this issue.
	\eat{The visual encoder and caption decoder are not tightly stitched together, but the previous methods made limited efforts to address this issue.}
	Consequently, the challenges are also presented in the language decoding phase. 
    The previous methods directly use the visual features as the initial visual ``word'' to generate the entire sentence, without considering the global semantics of the caption. This may mislead the language decoder to generate the sequence with generic content without details and highlights. For example, ``sunny day with blue sky'', ``red tennis court'' could be only described as ``sky'', ``tennis court''.

	\eat{In the RNN sentence decoder, the caption is predicted word-by-word based on every previous steps. During the decoder training, the gradient is back-propagated through RNN at word-level, the sentence-level global semantic features are weakly exploited. 
	As a result, the entire sentence is conditioned on the initial input, 
	making the initialisation extremely critical. 
	However, the previous methods directly use the visual features as the initial visual ``word''.
	
	The prediction is depended on the previous word, not the semantic of entire caption. The object relationships can be hardly explored, for example, the ``people ride a horse'' could be mistakenly described as ``people stand next to a horse''.}

	To address the challenges, we propose a novel modality transition image captioning framework to explicitly bridge the vision-language modality gap in Fig.~\ref{fig:overview_ours}. The framework is equipped with the novel Modality Transition Module (MTM), which is trained by the modality loss, and supported by our pre-trained text auto-encoder.
	In particular, the image is firstly forwarded into the CNN to obtain convolutional feature maps before passing to the region detectors (e.g. Faster-RCNN) for object-level features.
	Meanwhile, different from the previous baseline methods shown in Fig.~\ref{fig:overview_baseline}, the caption of the image is additionally encoded via the text auto-encoder. In this way, the global representation of the caption can be extracted.
	After both the image and caption representations are prepared, they are fed to the Modality Transition Module. In the MTM, the regional features are firstly average pooled, and passed through neural network layers to transfer into the preliminary textual vector. 
	The generated preliminary vector is compared with the global caption representation just encoded from the text auto-encoder by the modality loss, which measures the difference between the preliminary textual vector and the target global caption vector.
	It is important to let the model have the ability to generate the global textual features, because during testing, the caption and its global representation will not be available.
	The estimated global textual features are inputted to the language model to generate the sentence word-by-word. Optionally, during the word generating, visual attention can also be embedded to enhance the prediction performance by attentively shifting the sentence focus on different parts of the image.
	\eat{In the inference phase, the image is forwarded to the region detector to extract object features, and the visual features are transited into global textual representation via MTM, and finally the sentence is generated based on the global features by the language decoder.}

	\eat{Especially, in the image captioning task, the image-sentence pairs are given to supervise the image captioning model training.} 
	\eat{Before the captioning model training, we pre-train a dedicated text auto-encoder to encode the sentence into a global vector representation using only the sentences from the captioning corpus.}
	\eat{After the text auto-encoder is trained, we start to train the image captioning model with MTM.}
	The key contributions in this paper are three-fold:
	\vspace{-1ex}
	\begin{enumerate}
		\item To the best of our knowledge, it is the first work to propose a dedicated modality transition framework to explicitly bridge the gap between the visual and textual modalities in the image captioning model. 
		\item The proposed Modality Transition Module (MTM), modality loss, and the text auto-encoder are agnostic to the existing encoder-decoder framework, therefore, it is straightforward to embed this module into other models.
		\item Extensive quantitative and qualitative are conducted to evaluate the proposed framework comparing the the-state-of-arts methods, and different modality loss and base model structures are also studied to show the effectiveness of the proposed model.
	\end{enumerate}
	\vspace{-1ex}
	The rest of this paper is organised as follows: Sec.~\ref{sec:related_work} briefly reviews the literature in image captioning and multimodal fusion, Sec.~\ref{sec:method} demonstrates the proposed method, Sec.~\ref{sec:experiment} gives the experiments, and Sec.~\ref{sec:conclusion} concludes the paper.
	
	\section{Related Work} \label{sec:related_work}
	Visual captioning is an emerging knowledge discovery topic in conjunction with computer vision and natural language processing.
	The majority of existing models follow the encoder-decoder framework, in which the encoder is a visual feature extractor and the decoder is a language model. The captioning research has its origin in the early work on image captioning utilising the CNN as an image encoder, and the RNN as a language generation model~\cite{Karpathy_2017_tpami,Vinyals_2015_CVPR}. Similar frameworks are also proposed to accommodate paragraph generation~\cite{krause2016paragraphs,para_gan}, video captioning~\cite{video_caption_donahue,yang2018caption}.
	
	Furthermore, the visual attention mechanism is embedded into the visual captioning spectrum. During the sentence generation, the visual attention module is able to adaptively locate the most salient areas in the image given the current hidden states for more accurate prediction. Different structures are proposed such as soft attention~\cite{xu2015show}, adaptive attention~\cite{Lu_2017_CVPR}, depth attention~\cite{lookdeeper_mm18}, top-down and bottom-up attention~\cite{vqa_updown}, etc.
	
	Another line of work exploit reinforcement learning to optimise the model based on the evaluation metrics~\cite{rl_rennie,rl_Liu_2017_ICCV,curiosity_rl_luo}. These models take the action of predict the next word, given the states of visual, textual, and context vectors, to obtain the reward measured by caption evaluation scores.
	Other innovate methods focus on domain transfer~\cite{Chen_2017_ICCV,dual_cross_domain,multitask_cross_domain}, object hallucinations and gender bias in generated captions~\cite{womenSnowboard,objectHallucination}, topic modelling~\cite{topic_WangPYTM19}, Model efficiency~\cite{nips_transformer_objects, paic_adc2020, ijcai_cons_caption}, etc.
	
	Most of the existing work focuses on improving the vision or language feature extractions, and bridge them together by directly inserting the vision modalities in the language decoder. The modality gap is weakly addressed underlying the encoder-decoder framework, but improvements could still be made to explicitly engage the vision and language modality transitions.
	
	
	\section{Methodology}\label{sec:method}

	In this section, we demonstrate the proposed enhanced modality transition framework (Fig.~\ref{fig:overview_ours}) for image captioning. The novel framework consists of the text auto-encoder, the visual region detector, the modality transition module (MTM), the language decoder, and the visual attention. The details of each component in the framework are explained in this section.
	
	\subsection{Problem Formulation}
	The objective of image captioning is formulated in this section. The input RGB image is denoted as $I$. The output caption $\mathbf{S}$ consists of a sequence of words with length $N$. The final objective is to generate $\mathbf{S} = \{S_1, ..., S_N\}$ given $I$.
	
	\subsection{Text Auto-Encoder}
	In the training phase, the conventional image captioning model encodes $I$, but shelves the caption $\mathbf{S}$ only for calculating supervised loss to mimic the inference scenario where $\mathbf{S}$ is not available.
	However, both image $I$ and caption $\mathbf{S}$ are available during the training. The motivation is to investigate the informative caption during training for maximum utilisation in addition to be the ground-truth label. The text auto-encoder is proposed to encode
	information intensive captions into a global text representation, and guide the MTM model to transfer the visual features into the ``textual-like'' representations.
	
\begin{figure}[!b]
		\hspace{-0.5cm}
		\centering
		\subfloat[Text auto-encoder.\label{fig:ae_modules}]{%
			\includegraphics[trim=8.5cm 0cm 6cm 0cm, width=0.42\textwidth, clip=true]{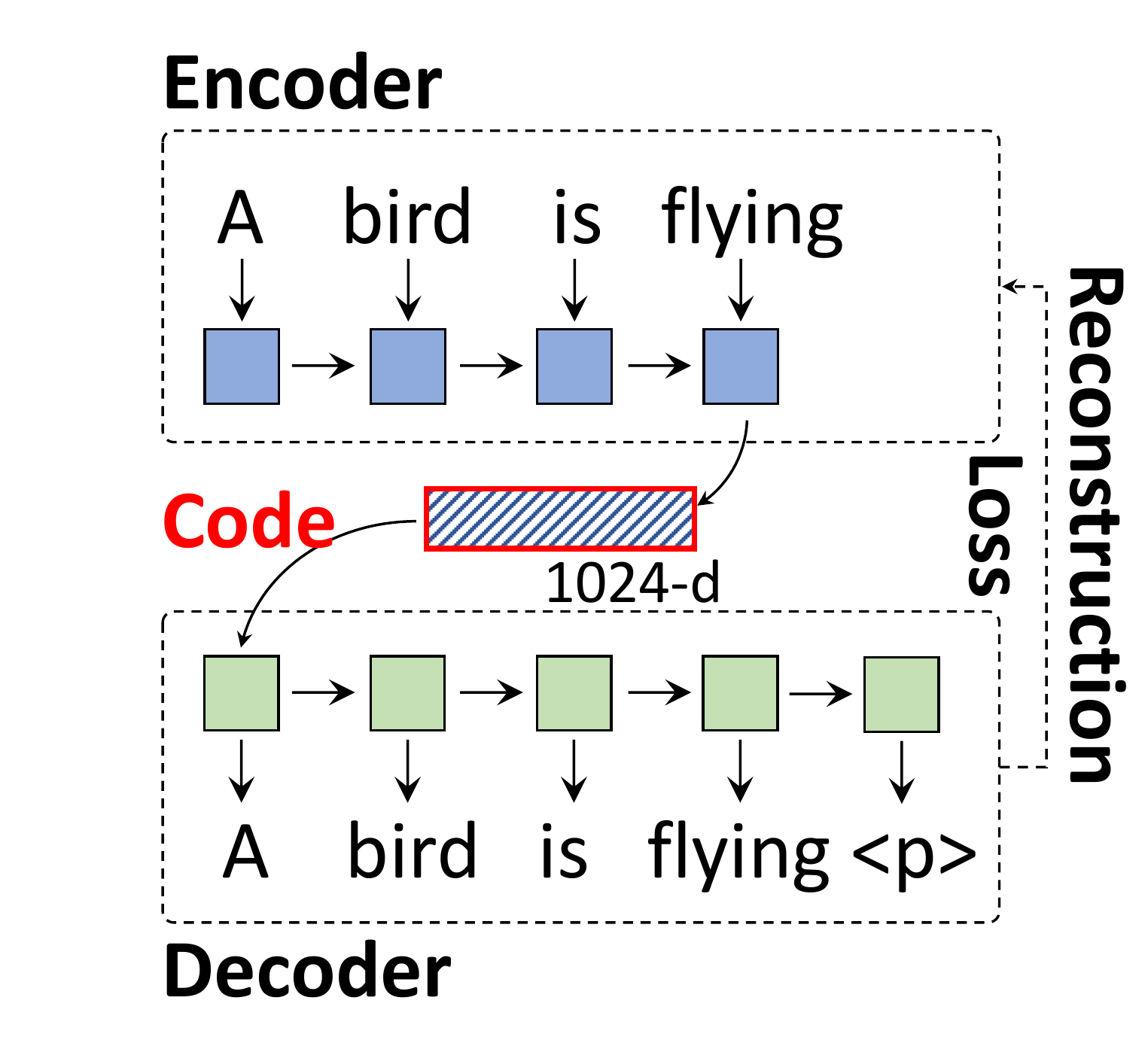}%
		}
		\hspace{0.5cm}
		\subfloat[Modality Transition Module.\label{fig:mt_modules}]{%
			\includegraphics[trim=6.5cm 1cm 6.3cm 1.5cm, width=0.55\textwidth, clip=true]{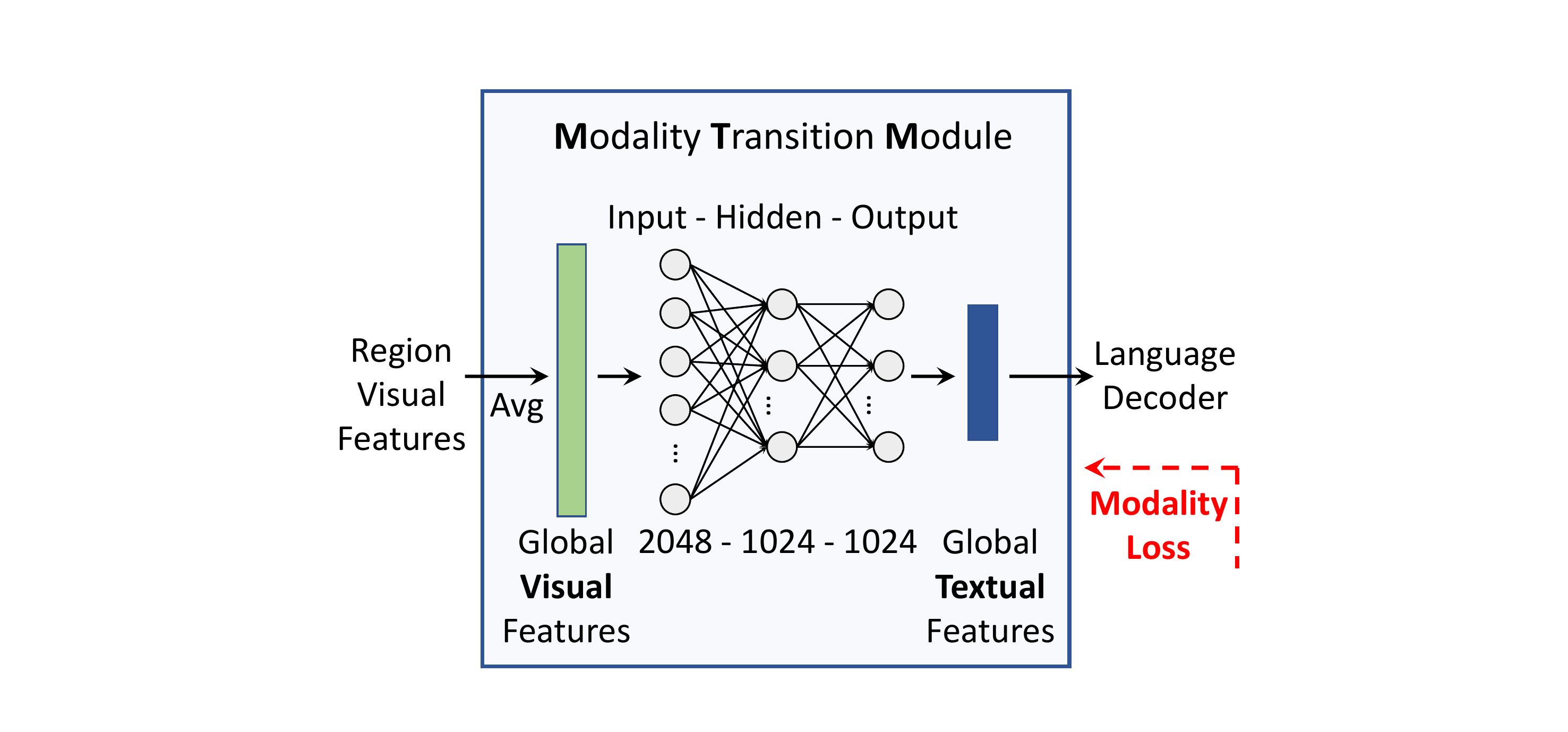}%
		}
		
		\caption{Modules of Modality Transition.}
		\label{fig:modules}
		
	\end{figure}

	The text auto-encoder consists of an encoder and decoder for automatically generating a compressed vector representation for the caption as illustrated in Fig.~\ref{fig:ae_modules}. The LSTM encoder-decoder adopts the sequence-to-sequence structure similar to neural machine translation~\cite{s2s_cho}. 
	In the text auto-encoder, the encoder module reads the input caption $\mathbf{S} = \{S_1, ..., S_N\}$, and encodes it to a fixed-length vector $u_g\in\mathbb{R}^{d_e}$. The LSTM is commonly used such that:
	\begin{align}
		h_t &= f_1(S_t, h_{t-1}, c_{t-1}),\\
		u_g &= h_N,
	\end{align}
	where $h_t \in \mathbb{R}^{d_e}$ is the hidden states at $t$, $c_{t-1} \in \mathbb{R}^{d_e}$ is the memory cell states at time step $t-1$, the dimension of encodings is $d_e$, $f_1$ is the LSTM, and $u_g$ is the generated global vector from the hidden states of the last time step $h_N$.
	
	Given the generated compressed caption code $u_g$ and all the predicted previous words $\{S_1, \cdots, S_{t'-1}\}$, the decoder LSTM is trained to predict the next word $S_{t'}$ in the caption. The probability over the entire caption is denoted as follows:
	\begin{equation}
		\log p(\mathbf{S}) = \sum_{t=1}^{N} \log p(S_{t'}|S_1, \cdots, S_{t'-1}, u_g), \label{eq:logprob}
	\end{equation} 
	where the dependency of model weights are dropped for convenience. We optimise the sum of log probabilities in Eq.~\ref{eq:logprob} among all the training samples using gradient descent. 
	Cross-entropy is utilised as the reconstruction loss to measure the difference between the re-generated and the original caption. By optimising all the parameters in both encoder-decoder
	, the auto-encoder is able to generate the compressed code $u_g$, which is able to reconstruct the original caption.
	
	\subsection{Modality Transition Module and Loss}
	
	\subsubsection{Modality Transition}
	The modality transition module projects the visual features into textual global vector for language decoder as illustrated in Fig.~\ref{fig:mt_modules}. With the modality transition module, the decoder is able to generate captions in an accurate and concise way.
	
	\noindent Before forwarding to MTM, the image $I$ is encoded via the region detector shown in Fig.~\ref{fig:overview_ours}, outputting visual features $\mathbf{V}= \{v_1, ..., v_K\} \in \mathbb{R}^{K \times d_v}$ with $K$ visual regions, where $d_v$ is the dimension of visual features.
	\begin{align}
		v_g &= \frac{1}{K}\sum_{i=1}^{K} v_i,\\
		u_g' &= \sigma (W_{2} \cdot (W_{1} v_{g} + b_1) + b_2),
	\end{align}
	where $v_g\in \mathbb{R}^{d_v}$ is the global (mean pooled) visual feature, $u_g'\in \mathbb{R}^{d_e}$ is the predicted global textual representations. $W_{1}\in\mathbb{R}^{d_e\times d_v}$, $W_{2}\in\mathbb{R}^{d_e\times d_e}$ denote the linear projection parameters and $b_1$, $b_2$ are the biases to be learned. $\sigma$ is the ReLU activation function.
	
	\subsubsection{Modality Loss}
	In the training phase, the predicted preliminary textual vector representation $u_g'$ is compared with the compressed code $u_g$ encoded by text auto-encoder. 
	The modality loss measures the difference between $u_g'$ and $u_g$, we empirically choose mean squared error (MSE) to implement the loss. Other distance measurements are also implemented and tested in Sec.~\ref{sec:loss}. The modality loss is written such that
	\begin{equation}
		\mathcal{L}_\mathcal{M}(u_g',u_g) = \frac{1}{d_e}\sum_{i=1}^{d_e} (u_{g'(i)} - u_{g(i)} )^2.
	\end{equation}
	
	\noindent We minimise the modality loss to guide the MTM to learn the correct projection from visual to textual vectors. This is necessary because there will be no caption during inference, so the real global textual vector will not be available.
	
	\subsection{Modality Transition-enhanced Captioner}
	Before the image captioning model training, the dedicated text auto-encoder is trained by reconstructing the caption itself to learn a compressed representation. During the image captioning model training, the total loss is the combination of cross-entropy loss $\mathcal{L}_\mathcal{E}$ and the proposed modality loss $\mathcal{L}_\mathcal{M}$. The total loss function is denoted as follows:
	\begin{equation}
		\mathcal{L} = \mathcal{L}_\mathcal{E} + \mathcal{L}_\mathcal{M}.
	\end{equation}
	The cross-entropy loss measures the performance of how well the predicted probability of the vocabulary reflects the ground-truth label. 
	
	\noindent\textbf{Summary}: To train the MTM, the outputs of MTM are compared with the auto-encoded reference global textual representations via modality loss. During the inference, as illustrated in Fig.~\ref{fig:overview_ours}, the visual features are firstly extracted by the region detector such as Faster-RCNN. Secondly, before passing the visual features to the language decoder, Modality Transition Module (MTM) projects the averaged visual features into global textual representations. The language decoder decodes the representations into captions with the optional visual attention module such as the TopDown attention~\cite{vqa_updown}.

	\section{Experiments} \label{sec:experiment}
	\subsection{Experimental Settings}
	\subsubsection{Dataset}
		The experiments are conducted on the MS-COCO~\cite{lin2014microsoft} Captioning dataset for evaluating the proposed MTM model. The train, validation, and test partitions are following the commonly adopted ``Karpathy''~\cite{Karpathy_2017_tpami} split, which has 113,287 training, 5,000 validation, and 5,000 test images.
	\subsubsection{Implementation Details}
		The proposed framework consists of the visual region detector, the MTM module, the text auto-encoder, the visual attention, and the language decoder.
		For visual region detector, we extract the region features using Faster-RCNN with ResNet-101 backbone following~\cite{vqa_updown}.
		In the MTM module, the model has two fully-connect neural network layers with 2048-1024 and 1024-1024 dimensions for each layer, i.e., $d_v=2048$, $d_e=1024$.
		The text auto-encoder consists of an encoder and decoder, both of them are implemented with 2-layer LSTM with 1024 dimensions.
		The captioning model is optimised by ADAM optimiser~\cite{adam_KingmaB14}, and the learning rate is empirically set to $5e-4$ with decay.
		All the models in this paper are trained on a server with GeForce GTX 1080Ti GPU.

	\subsubsection{Evaluation Metrics}
		For comparison, we report the performance measured by automatic language evaluation metrics: BLEU-$n$~\cite{papineni2002bleu}, METEOR~\cite{lavie2005meteor}, Rouge-L~\cite{lin-2004-rouge} and CIDEr~\cite{vedantam2015cider}. For example, the BLEU-$n$ measures $n$-gram precision of the candidate caption.
		\eat{The BLEU-$n$ measures $n$-gram precision of the candidate caption, METEOR also considers recall in addition to precision, Rouge-L focuses on the longest co-occurring in sequence $n$-grams in similarity calculation, and CIDEr utilises TF-IDF in cosine similarity measurement with scaling factor.}
		
	\subsubsection{Compared Methods} \label{sec:compared_methods}
		We compare the proposed method with a number of encoder-decoder image captioning models. \textbf{ShowTell}~\cite{Vinyals_2015_CVPR}: The neural image captioner consists of CNN image encoder and LSTM language decoder. \textbf{Adaptive}~\cite{Lu_2017_CVPR}: The CNN-LSTM captioner with the adaptive visual attention. \textbf{Att2in} and \textbf{Att2all}: Variants of attention-based models trained with self-critical sequence training (SCST) from~\cite{rl_rennie}. \textbf{UpDown}: state-of-the-art captioning model with carefully engineered bottom-up and top-down attention~\cite{vqa_updown}.
		In addition, in the following comparison tables and figures, suffix \textbf{MT} indicates that the model has Modality Transition Module embedded, and suffix \textbf{RL} indicates that the model is fine-tuned by reinforcement learning objectives from the SCST~\cite{rl_rennie}.
		
	\subsection{Quantitative Analysis}
	We embedded the proposed MTM model to the simplest baseline ShowTell and state-of-the-art model UpDown, and labelled as \textbf{ShowTell\_MT} and \textbf{UpDown\_MT}, respectively in Table~\ref{tab:comparison_offline}. The reinforcement learning objective is also implemented to achieve the best performance.
	From Table~\ref{tab:comparison_offline}, by exploiting the MTM module, we can clearly observe that the best model UpDown\_MT\_RL can improve the previous state-of-the-art model UpDown\_RL by relative 3.4\% in CIDEr. Consistently, in terms of the baseline ShowTell model, when the MTM is deployed, the CIDEr performance increases from 96.29 to 100.42. Moreover, when self-critical reinforcement learning is adopted to optimise ShowTell\_MT, the BLEU-4 is boosted by 10.5\%, and the CIDEr result is increased from 100.42 to 113.37. Importantly, the ShowTell\_MT\_RL is not equipped any form of attention mechanism, but the performance is surpassing attention-based models such as Adaptive and Att2in\_RL and comparable to Att2all\_RL. 
	\eat{The quantitative analysis demonstrates that the proposed method can improve the image captioning 
	performance by explicitly bridging the vision and language modalities through the modality transition.}
	Comparing to the previous work, where the visual features are assumed as the modality-invariant global representations for language decoding without compensation, the MTM model significantly improves language generation quality by allowing the smooth transition from vision to language.

	\begin{table}[!t]
		\centering
		\caption{Performance comparison on MSCOCO Karpathy test split \cite{Karpathy_2017_tpami}. The results of the proposed MTM are annotated with \textbf{MT}, and the self-critical optimised results are end with \textbf{RL}.}
		\label{tab:comparison_offline}
		\begin{tabular}{p{3cm}*{6}{c}}
			\toprule
			Model & BLEU-3         & BLEU-4         & METEOR         & Rouge-L        & CIDEr           \\
			\midrule
			ShowTell~\cite{Vinyals_2015_CVPR}           & 41.60           & 30.34          & 25.05          & 53.58          & 96.29           \\
			Adaptive~\cite{Lu_2017_CVPR} & 43.90          &    33.20      & 26.60          &  -         &   108.50        \\
			Att2in\_RL~\cite{rl_rennie} &  -        &  33.30       &  26.30      &  55.30     &  111.40      \\
			Att2all\_RL~\cite{rl_rennie} & -  &  34.20     &  26.70      &  55.70     &  114.00   \\
			UpDown~\cite{vqa_updown} & - & 36.20        & 27.00         & 56.40   & 113.50\\
			UpDown\_RL~\cite{vqa_updown} & - & 36.30 & 27.70 & 56.90 & 120.10\\
			\midrule
			
			\textbf{ShowTell\_MT} & 42.05 & 30.97 & 25.67 & 53.77 & 100.42\\
			\textbf{ShowTell\_MT\_RL} & 45.91 & 34.21 & 26.58 & 55.96 & 113.37\\ 
			\textbf{UpDown\_MT} & 47.08 & 36.52 & 27.67 & 56.59 & 113.78\\
			\textbf{UpDown\_MT\_RL} & \textbf{49.41} & \textbf{37.50 }& \textbf{28.23} & \textbf{57.88} & \textbf{124.14}\\ 
			
			\bottomrule
		\end{tabular}

	\end{table}

	\subsection{Qualitative Analysis}
\begin{figure}[!t]
			\centering
			\includegraphics[trim=0cm 0cm 0cm 0cm, width=0.98\textwidth]{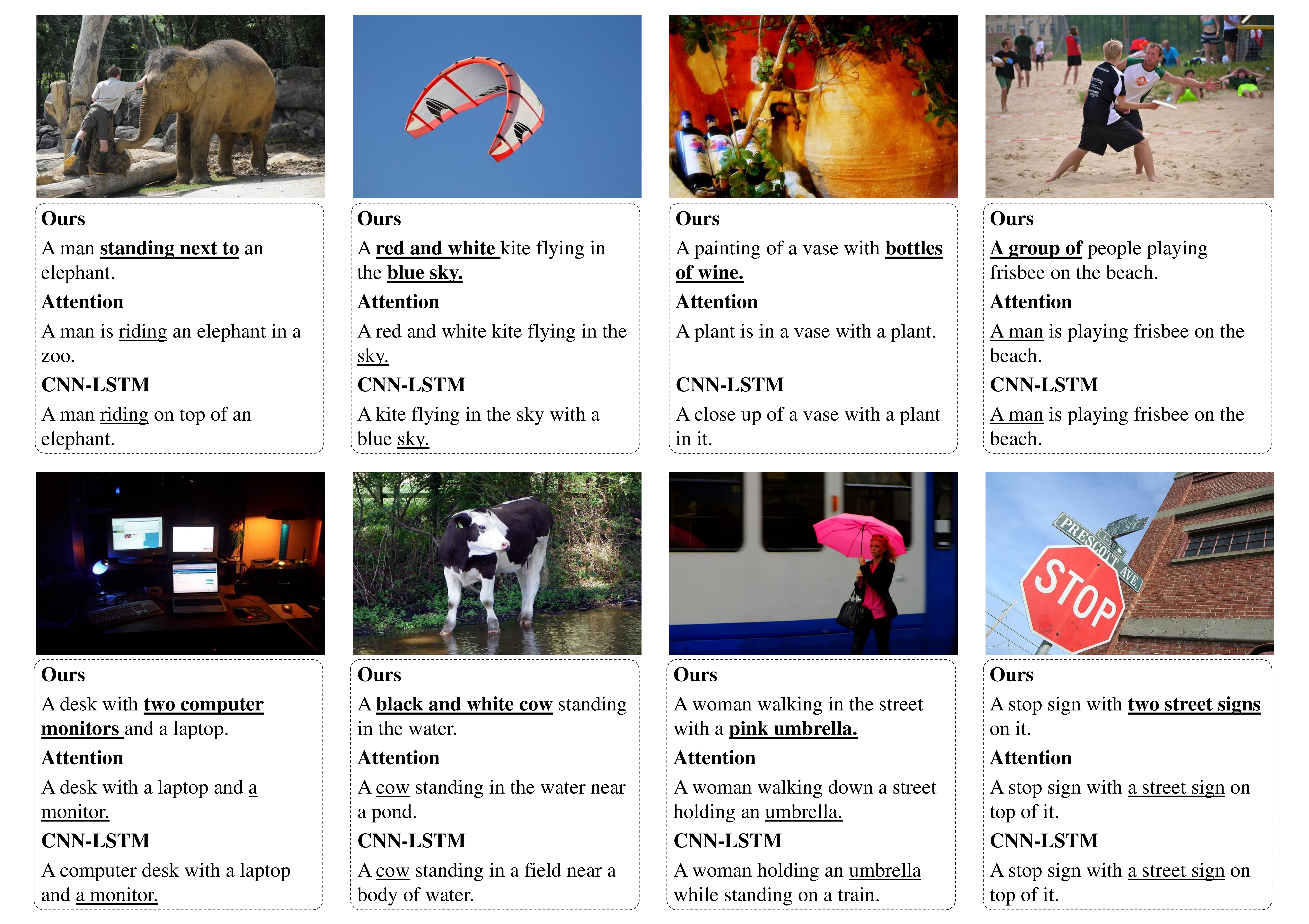}
			\caption{Case studies of MTM, Attention model, and basic CNN-LSTM model.}
			\label{fig:case_examples}
		\end{figure}

		In qualitative study, we present a number of showcases to demonstrate the performance of MTM intuitively. Comparing to the results from UpDown attention~\cite{vqa_updown} model and ShowTell CNN-LSTM~\cite{Vinyals_2015_CVPR} model shown in Fig.~\ref{fig:case_examples}, the proposed MTM is able to generate sentences in a accurate, detailed and grammatically correct way. 
		Specifically, after modality transition, the language model can precisely identify object-interactions from the scene. For example, in the first row, the first example illustrates that the MTM can detect the fact that a man is ``standing next to'' rather then ``riding'' the elephant. The ability of differentiate human-object interactions also is also shown in the 4th image of the top row. Although only one man is holding the frisbee in that picture, from the action and pose we can conclude that a group of people are playing rather than only one people. 
		Furthermore, the details can be better preserved by exploiting global caption representations in MTM. The pictures on the 2nd column give accurate colours of the objects, while the 3rd image on the top and the 1st on the bottom reveal details such as ``bottles of wine'', ``two computer monitors'' rather than the general expressions from other models.
		
\begin{figure}[!b]
	\begin{floatrow}
		\centering
		\capbtabbox{%
			\centering
			\hspace{-0.2cm}
			\begin{tabular}{p{2.3cm}p{1.0cm}p{1.0cm}p{1cm}}
				\toprule
				Model 		& B4 & M & C \\
				\midrule
				
				ShowTell (Base)           & 30.34          & 25.05          & 96.29           \\
				\midrule
				KLD  & 11.92 & 14.46 & 28.11 \\ 
				COS & 13.00 & 15.26 & 31.83 \\ 
				MMD & 30.28 & 25.05 & 96.27\\ 
				MAE & 30.77 & 25.44 & 98.95\\ 
				\textbf{MSE (Ours)} & \textbf{30.97} &\textbf{25.67} & \textbf{100.42}\\ 
				\bottomrule
			\end{tabular}
		}{%
			\caption{Comparison for different variants of modality loss.}\label{tab: variants}
		}
		\hspace{0cm}
		\ffigbox{%
			\centering
			\hspace{-0.1cm}
			\includegraphics[trim=0cm 0.3cm 0cm 0cm, width=0.44\textwidth, height=3.7cm]{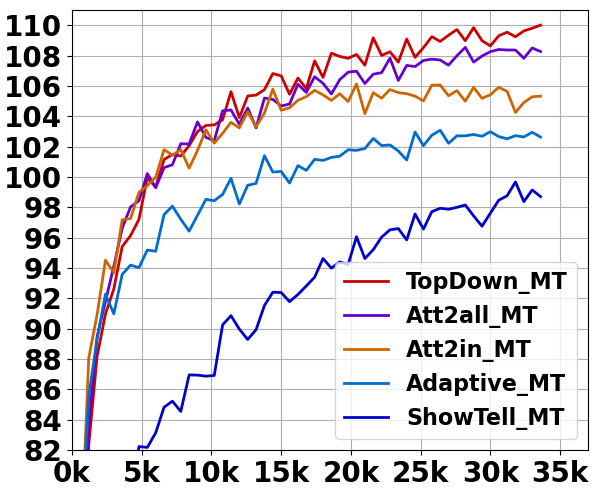}
		}{%
			\caption{Training steps and performances of CIDEr.}
			\label{fig:model_train}
		}
	\end{floatrow}
\end{figure}	

	\subsection{Ablation Study}
	\subsubsection{Modality Loss} \label{sec:loss}
		In this section, we compare different variants of modality loss.
		In Table~\ref{tab: variants}, ShowTell indicates the performance of the base model without modality transition. All other models in the table are implemented with MTM.
		For comparison, we choose different measurements between the predicted and the target textual encodings, the loss functions include mean absolute error (MAE), mean squared error (MSE), Cosine distance (COS), Kullback-Leibler divergence (KLD), Maximum Mean Discrepancy (MMD)~\cite{mmd}. From the comparison result, we can see that the standard MSE effectively measures the loss between the preliminary textual vectors and the well-trained global caption representations giving the best performance. However, the measurements such as KLD and COS result in poor performance.
		The KLD measures the relative entropy between two distributions, but not directly computing the distances between the input and target vectors. While the COS measures the angles between the two vectors, but the embedding vectors are sensible to magnitude.
		The experiments show that they are not suitable for the modality similarity measures.

\begin{figure}[!t]
	\centering
	\includegraphics[trim=6cm 0cm 6cm 0cm, width=0.90\textwidth, height=4cm]{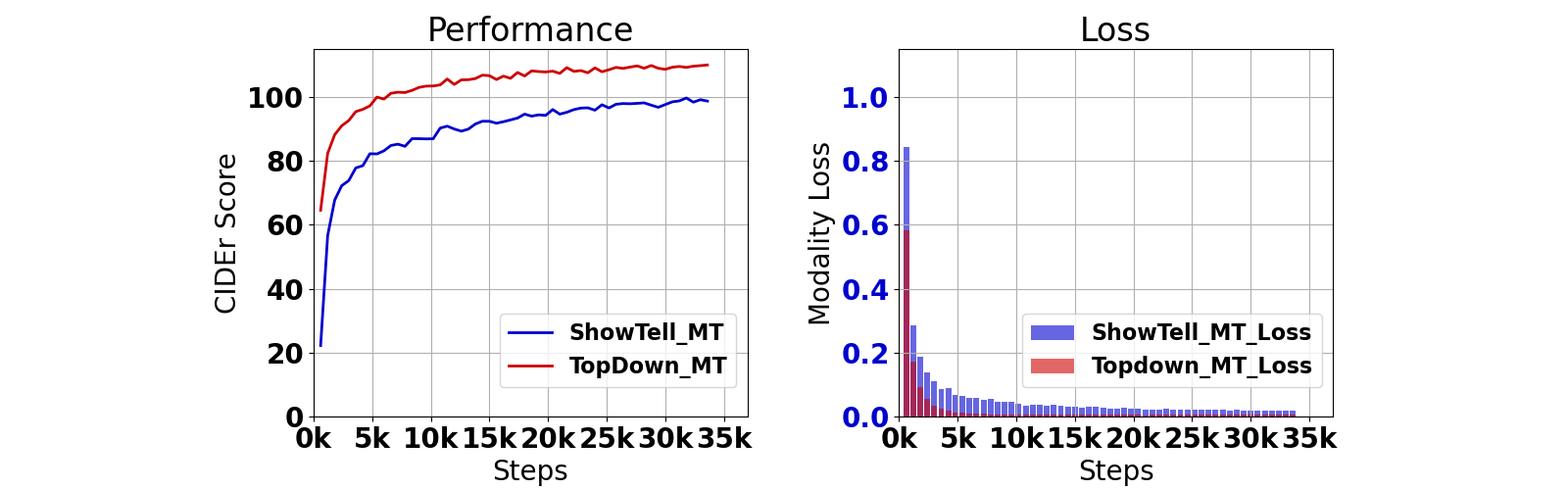}
	\caption{Training modality loss and performance gain.}
	\label{fig:score_loss_combo}
\end{figure}

	\subsubsection{Modality Transition Module Training}

		We implement the Modality Transition Module on the existing models mentioned above in Sec.~\ref{sec:compared_methods}, and annotated with suffix \textbf{MT}: ShowTell\_MT, Adaptive\_MT, Att2in\_MT, Att2all\_MT, and TopDown\_MT. The detailed training curves are illustrated in Fig.~\ref{fig:model_train}, and the steadily increasing red line on the top is the proposed MTM built on TopDown~\cite{vqa_updown} attention.
		MTM also performs well on other encoder-decoder models, demonstrating strong ability to be embedded in the conventional image captioning models with good adaptability.
		
		As an auxiliary study, the validation performance and the training modality loss during model learning are visualised in the Fig.~\ref{fig:score_loss_combo}. The two methods included are MTM modules embedded on ShowTell and TopDown. As shown in the figure, with the CIDEr scores increasing on the left, the corresponding modality losses are decreasing in the similar pace on the right sub-figure. We can also observe that the TopDown\_MT performs better than ShowTell\_MT in terms of CIDEr scores, and the TopDown\_MT's loss is lower as expected. This indicates that the proposed modality loss is strongly correlated to the model performance.

	\section{Conclusion}\label{sec:conclusion}
		In this work, we propose a Modality Transition Module (MTM) to enhance vision and language modalities transition for image captioning. The proposed model explicitly projects the image visual features into the global textual representation vector, giving the language decoder the preferable textual cues for caption generation. The experiments demonstrate the effectiveness of the proposed model.

	\section*{Acknowledgement}
	This work is partially supported by ARC DP190102353 and ARC DP170103954.
	
	%
	%
	%
	%
	\bibliographystyle{splncs04}
	\bibliography{bib_caption}
	
\end{document}